%% file: main_ai4cc.tex
\documentclass[10pt,twocolumn,letterpaper]{article}

\usepackage{cvpr}              %

\usepackage{graphicx}
\usepackage{amsmath}
\usepackage{amssymb}
\usepackage{booktabs}

\usepackage[pagebackref,breaklinks,colorlinks]{hyperref}

\usepackage[capitalize]{cleveref}
\crefname{section}{Sec.}{Secs.}
\Crefname{section}{Section}{Sections}
\Crefname{table}{Table}{Tables}
\crefname{table}{Tab.}{Tabs.}

\newcommand{\figref}[1]{Fig. \ref{#1}}
\newcommand{\tabref}[1]{Tab. \ref{#1}}

\newcommand{\secref}[1]{Sec. \ref{#1}}

\usepackage{algorithm}
\usepackage{algorithmic}
\usepackage{mathtools}
\usepackage{amsthm}
\usepackage{graphbox}
\usepackage{xcolor}
\usepackage{soulpos}
\usepackage{multirow}
\usepackage[capitalize]{cleveref}
\crefname{section}{Sec.}{Secs.}
\Crefname{section}{Section}{Sections}
\Crefname{table}{Table}{Tables}
\crefname{table}{Tab.}{Tabs.}

\definecolor{swcolor}{rgb}{0.607,0.149,0.713}

\definecolor{todocolor}{rgb}{0.5, 0.0, 1.0}

\newcommand{\rom}[1]{\uppercase\expandafter{\romannumeral #1\relax}}

\newcommand\blfootnote[1]{%
  \begingroup
  \renewcommand\thefootnote{}\footnote{#1}%
  \addtocounter{footnote}{-1}%
  \endgroup
}

\newcommand*\samethanks[1][\value{footnote}]{\footnotemark[#1]}
\def\cvprPaperID{13} %
\def\confName{CVPR}
\def\confYear{2023}

\begin{document}

\title{User-friendly Image Editing with Minimal Text Input: Leveraging Captioning and Injection Techniques}

\author {
    Sunwoo Kim\thanks{Equal contribution} \,\textsuperscript{\rm 1},
    Wooseok Jang\textsuperscript{\rm 1},
    Hyunsu Kim\textsuperscript{\rm 2},
    Junho Kim\textsuperscript{\rm 2}, \\
    Yunjey Choi\textsuperscript{\rm 2}, 
    Seungryong Kim\thanks{Corresponding author} \,\textsuperscript{\rm 1}, 
    Gayeong Lee\samethanks\, \textsuperscript{\rm 2} \\ \\
    \textsuperscript{\rm 1} Korea University, South Korea \hspace{5pt}
    \textsuperscript{\rm 2} NAVER AI Lab \hspace{5pt} \\
     {\tt\small \textsuperscript{\rm 1}\{sw-kim, jws1997, seungryong\_kim\}@korea.ac.kr } \\[-3pt]
    {\tt\small \textsuperscript{\rm 2}\{hyunsu1125.kim, jhkim.ai, yunjey.choi, gayoung.lee\}@navercorp.com}
    }

\input{title.tex}
\blfootnote{${}^{*}$Equal contribution}
\blfootnote{${}^{\dagger}$Corresponding author}

\begin{abstract}
Recent text-driven image editing in diffusion models has shown remarkable success. However, the existing methods assume that the user's description sufficiently grounds the contexts in the source image, such as objects, background, style, and their relations. This assumption is unsuitable for real-world applications because users have to manually engineer text prompts to find optimal descriptions for different images.
From the users' standpoint, prompt engineering is a labor-intensive process, and users prefer to provide a target word for editing instead of a full sentence.
To address this problem, we first demonstrate the importance of a detailed text description of the source image, by dividing prompts into three categories based on the level of semantic details. Then, we propose simple yet effective methods by combining prompt generation frameworks, thereby making the prompt engineering process more user-friendly. Extensive qualitative and quantitative experiments demonstrate the importance of prompts in text-driven image editing and our method is comparable to ground-truth prompts.

\end{abstract}

\vspace{-8pt}
\section{Introduction}
\label{intro}

Diffusion-based large-scale text-conditional image synthesis frameworks~\cite{nichol2021glide, rombach2021highresolution, ramesh2021zero,  saharia2022photorealistic} have shown impressive generation performance and applications. Among the various applications, text-guided image editing has received significant interest for its user-friendliness and quality~\cite{avrahami2022blendedlatent, hertz2022prompt, Kawar2022ImagicTR, wu2022uncovering, brooks2022instructpix2pix, parmar2023zero}.
Although the importance of text prompts in image generation is well-known, their role in text-guided image editing, which conditions the image to be generated following the meaning of text prompts has not been thoroughly analyzed in previous literature. For instance, when we edit a cat in the source image to the dog, text prompts describing the source image can vary; from one noun (\eg, ``cat'') to a full-sentence description (\eg, ``a cat sitting next to a mirror'') more grounded to the entire image content. 

In this paper, we categorize prompts into three levels according to description details and evaluate their impacts on editing quality in standard benchmarks~\cite{radford2021learning, zhang2018perceptual}.
Simple text prompts degrade the editing performance in terms of localization; unwanted regions of the source image are edited.
While text tokens in a detailed text prompt usually direct attention to semantically relevant regions, text tokens in a simple text prompt may not be sufficient for accurate localization, as they may be associated with non-correlated regions.
This problem arises due to missing contexts in the input prompt about the source image. 
Our findings suggest that a detailed text prompt plays a crucial role in text-guided image editing. Specifically, providing detailed prompts leads to a significant improvement in editing performance. These results suggest that users should engage in prompt engineering to find detailed text prompts to yield the best editing quality. 

However, fully describing an image with detailed prompts is time-consuming and error-prone in real-world editing scenarios. To address this issue, we propose to utilize the prompt generation models~\cite{wen2023hard, li2022blip} to bridge the gap between using a simple prompt by the user and a fully described prompt. One model~\cite{li2022blip} utilizes a captioning model to automatically generate prompts for missing contextual information, which is not described in the given prompts. The other model~\cite{wen2023hard} employs \textit{hard prompts} by text prompt optimization.  
We further propose methods for integrating user-provided source attributes, which specify the regions to be edited in the source image, and filtering out inaccurate tokens in these models. We demonstrate the effectiveness of our methods on a variety of editing tasks using both real and synthetic images. Our proposed methods enable more intuitive image editing based on human instructions.

\section{Related work}
\paragraph{Image editing with diffusion models.}
The task of generating images conditioned on corresponding text, known as text-to-image synthesis, has recently gained attention thanks to Diffusion Probabilistic Models~\cite{ho2020denoising}. Several diffusion-based large-scale text-to-image synthesis frameworks have been proposed, including OpenAI's DALLE-2~\cite{ramesh2021zero}, Google's Imagen~\cite{saharia2022photorealistic}, and Stability AI's Stable Diffusion~\cite{rombach2021highresolution}. These models are capable of synthesizing any type of image with high fidelity and diversity, which has led to recent efforts to explore text-guided single-image editing.

One such effort is SDEdit~\cite{meng2021sdedit}, which proposes a simple method of adding noise to the input image and then denoising it with a corresponding text condition following a predefined diffusion schedule. Other methods~\cite{nichol2021glide, avrahami2022blended, avrahami2022blendedlatent} exploit masks from user input to constrain the regions to be edited. However, these mask-based approaches require a tedious masking process, which can be bothersome to users.
Despite these advancements, simply adding Gaussian noise and denoising the corrupted image may result in undesired editing effects in the unmasked regions. To address this, some recent works, such as DiffEdit~\cite{couairon2022diffedit} and Prompt-to-Prompt~\cite{hertz2022prompt}, adopt the DDIM inversion~\cite{song2020denoising} method and propose methods for automatically generating masks.
Furthermore, Prompt-to-Prompt manipulates cross-attention layers for better structure preservation. However, these approaches are still dependent on the users' input prompts and do not fully consider the strong influence of input prompts on the editing results.

\vspace{-10pt}
\paragraph{Finding text condition in diffusion model.}
Recently, Gal et al.~\cite{gal2022image} and Ruize et al.~\cite{ruiz2022dreambooth} suggest a textual inversion scheme for diffusion models that enables regenerating a user-provided concept out of 3-5 images. However, these models are designed for the object centered in the image, rather than for general objects existing anywhere in the scene.
For the general purpose, Mokady et al.~\cite{mokady2022null} iteratively optimize the null-text embeddings to invert an input image with the user-given text prompt.
Although such an optimization technique improves the reconstruction quality, the editing results are still dependent on the input prompt.

To alleviate the burden of prompt engineering by users, Brooks et al.~\cite{brooks2022instructpix2pix} utilizes GPT-3~\cite{brown2020language} to augment text captions and proposed an instructive text-to-image editing framework. However, such an approach needs to do manual labeling and collect lots of data. Parmar et al.~\cite{parmar2023zero} propose a training-free framework, called pix2pix-zero. They find the editing direction using a bank with hundreds of texts of diverse sentences from GPT-3~\cite{brown2020language}. But, when the targeted attributes are not captured in the generated prompts from BLIP~\cite{li2022blip}, the embedding is mapped to the unintended point, which outputs undesirable editing results.

Other approaches~\cite{mokady2022null, wu2022uncovering} have shown the feasibility of using prompts generated from off-the-shelf learning-based~\cite{brown2020language} and optimization-based~\cite{wen2023hard} frameworks. To avoid labor-intensive processes such as prompt engineering and generating numerous text prompts, we utilize an off-the-shelf captioning model to obtain a prompt that sufficiently describes the context of the entire scene in the input image. Additionally, we propose a technique for injecting target editing words to address the limitation of pix2pix-zero~\cite{parmar2023zero}. Our approach mitigates the dependence on captioning models such as BLIP~\cite{li2022blip}.

\vspace{-5pt}
\section{Analysis of text prompts in diffusion-based image editing}%
\label{analysis}

\begin{figure}[t]
    \centering
    \includegraphics[width=1.0\linewidth]{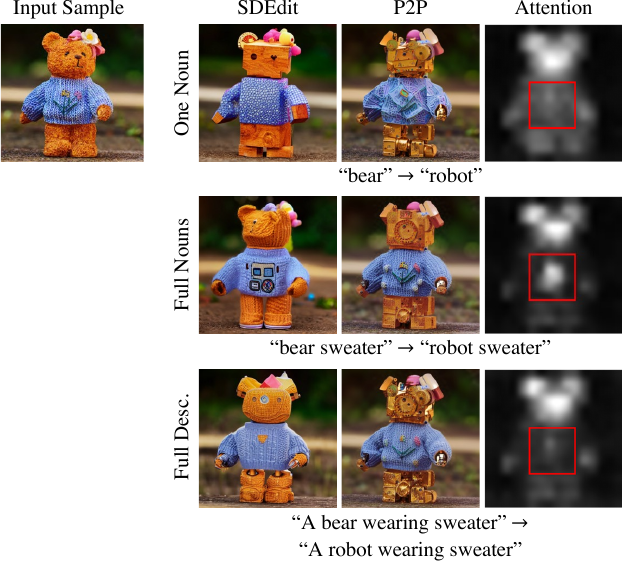}
    \vspace{-17pt}
    \caption{\textbf{Emerging problems from lack of grounding in the prompts.} When the details of the prompts lack, problems are observed, such as context disappearance and editing unintended regions. The first column shows the results from the SDEdit~\cite{meng2021sdedit}, the second column shows the results of Prompt-to-Prompt~\cite{hertz2022prompt}, and the third column shows the visualization of values in cross-attention activated by text token of ``bear''.}
    \label{fig:problem} 
\end{figure}

In this section, we analyze the influence of text prompts on the three text-guided editing frameworks, including SDEdit~\cite{meng2021sdedit}, Prompt-to-Prompt~\cite{hertz2022prompt} and Null-text Inversion~\cite{mokady2022null}. We use the Stable Diffusion~\cite{rombach2021highresolution} as the backbone for text-guided generation frameworks. In the problem setting, given a source image $\mathcal{I}$, we aim to edit $\mathcal{I}$ following the text guidance from an edited prompt. The edited prompt $\mathcal{P}^*$ is defined by replacing some tokens in the source prompt $\mathcal{P}$. The prompt $\mathcal{P}$ is encoded through the CLIP's text encoder $\psi_{T}$ to guide the latent feature in the diffusion sampling process. For example, as exemplified in the~\figref{fig:problem}, when we give a source prompt "A bear wearing..." which describes the source image, we change the bear with a robot by the edited prompt "A robot wearing...".

\subsection{Categorize text prompts by details}
We first investigate the influence of the source prompt $\mathcal{P}$ and edited prompt $\mathcal{P}^*$ on the quality of edited images in diffusion-based text-driven editing frameworks~\cite{meng2021sdedit,hertz2022prompt,mokady2022null}. We divide kinds of the prompt into three levels according to the amount of information the prompt has: (i) \emph{One Noun} (e.g., ``a cat'' $\rightarrow$ ``a dog''), where users only provide a word to identify the parts to be edited in the source and target images, respectively; (ii) \emph{Full Nouns} (e.g., ``clam pasta dish'' $\rightarrow$ ``shrimp pasta dish''), where users provide a pair of words as well as other words to describe the parts that should not be changed after the editing operation; and (iii) \emph{Full Description} (e.g., ``A clam pasta on the dish'' $\rightarrow$ ``A shrimp pasta on the dish''), where users provide a detailed text condition in the form of a sentence that contains more information such as verb than using only words for conditioning.

\subsection{Effect of text prompt details}
\figref{fig:problem} illustrates how editing quality can vary depending on prompt details. In the example shown, the user wants to change ``a bear wearing a sweater'' to ``a robot wearing a sweater''. In the \emph{One Noun} setting, where the prompt only includes the word ``robot'' to designate the edited part, the model also changes other attributes (the sweater) of the image, indicating that the lack of information about the specific part to be edited results in inaccurate attention maps. In contrast, in the \emph{Full Nouns} setting, the prompt includes additional information about the attribute to be preserved (the sweater), so the model is able to maintain the sweater while editing the bear to a robot. In the \emph{Full Description} setting, where the prompt includes even more detailed information like verbs, the model achieves finer-grained localization and preserves the pattern of the sweater during the editing process. In the third column in ~\figref{fig:problem}, we can observe that as the semantic information in the prompt decreases, the attention maps become less specific and generate more widespread attention, which can lead to less accurate and precise image editing results. In other words, a detailed prompt is critical for achieving accurate and high-quality image editing, as it helps the cross-attention layer to localize the region that needs to be modified accurately. 

To quantitatively measure the quality of editing based on the level of prompt detail, we measured the CLIP score~\cite{radford2021learning} and LPIPS~\cite{zhang2018perceptual} that indicate faithfulness to the corresponding texts and preservation capability about other details except for candidate parts to be edited. The CLIP score shows how well the target is reflected in the edited image, while LPIPS indicates how well the original image is preserved. Therefore, there is a trade-off between the two metrics, so we presented it as a graph in ~\cref {fig:quan_real}. The detailed evaluation protocol will be presented in ~\secref{sec:experiments}. We can see that the editing results become better quantitatively with more detailed prompts.

Based on these observations, we argue that in text-driven editing models~\cite{meng2021sdedit,hertz2022prompt} that rely on attention maps generated from prompts without fine-tuning, the quality of the attention maps becomes crucial for achieving high-quality edits. However, for users, generating detailed text descriptions every time they want to do image editing can be very cumbersome. In the next section, we propose a solution to address the lack of the semantic problem in the prompt by leveraging a pre-trained model to fill in missing information when users provide minimal words.

\begin{figure}[t]
\centering
\includegraphics[width=1.\linewidth]{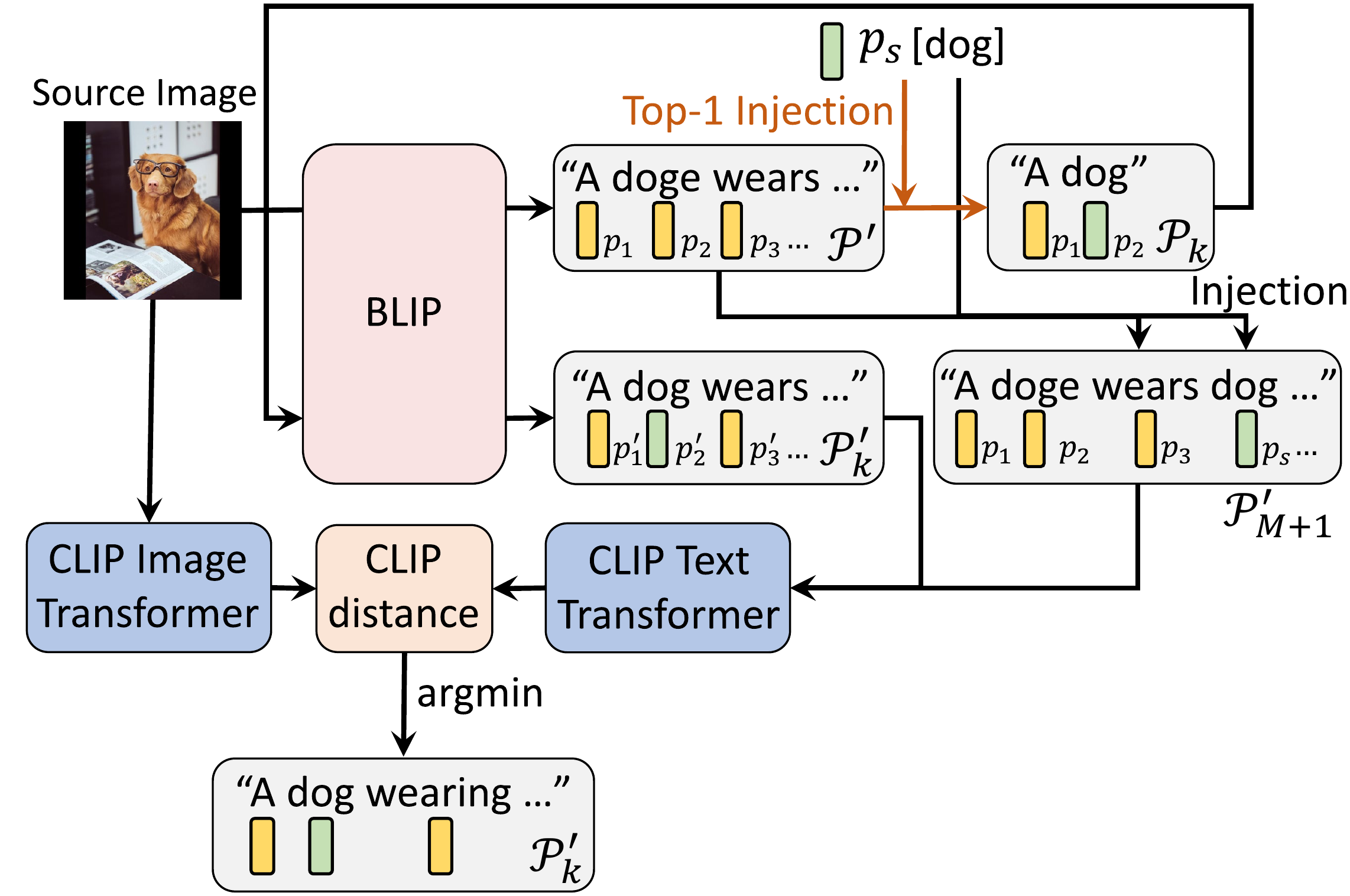}\\
\vspace{-5pt}
    \caption{\textbf{Illustration of captioning-based method.}   
    }
    \label{fig:captioning_method}
\end{figure}

\section{Proposed method}
In the analysis from~\secref{analysis}, we observe that the editing frameworks~\cite{meng2021sdedit,hertz2022prompt,mokady2022null} generate degraded results where only conditioned on the prompts of the \emph{One Noun} setting. However, in real-world applications, it is more intuitive and user-friendly for users to provide only the attributes they want to change rather than full descriptions, such as "a cat" $\rightarrow$ "a dog" and "blue hair" $\rightarrow$ "black hair".  To mitigate this problem, we propose to utilize prompt generation frameworks~\cite{wen2023hard, li2022blip} to complement the lack of other contextual information, but the generated prompts by captioning model itself are often unsuitable for directly utilizing in image editing, as these prompts may not include text tokens of source attributes to be edited and sometimes redundant tokens are included.

To alleviate this, we propose a simple yet effective solution to inject source attributes into generated prompts utilizing the captioning model~\cite{li2022blip} and hard prompt optimization~\cite{wen2023hard} and a way of removing the redundant tokens. To inject the source attributes into the generated prompts, we need to consider the case where a synonym for the source attribute exists within the generated prompts or does not exist. In the former case, we should think about the location where the source attributes will be injected. In the letter case, we have to consider how we address the synonym for the source attributes which exist within the generated prompts. In the following, we explain these methods in detail.

\vspace{3pt}
\subsection{Captioning-based method}
In this section, we first propose a captioning-based method. We use BLIP~\cite{li2022blip}, vision-language pre-training with image transformer ViT~\cite{dosovitskiy2020vit}, denoted by $E_{I}$, which is initialized pre-trained on Imagenet, and the text transformer initialized from $\text{BERT}_\text{base}$~\cite{devlin2018bert} with $14M$ images in total. Then captioner is fine-tuned with a language modeling loss to generate captions given images. After finetuning, we can autoregressively generate the caption with image embedding $E_{I}(\mathcal{I})$ of the source image:
\begin{equation}
p_{M} = LM(p_1, \ldots, p_{M-1} ; E_{I}(\mathcal{I}))
\end{equation}
where $M$ is the number of tokens and $[p_1, \ldots, p_{M}]$ are the sequence of text tokens of generated caption $\mathcal{P}$. 

However, it is not assured that the source attributes that we want to edit are not always captured in the generated prompts $\mathcal{P'}$. For image editing, we need to replace source attributes $p_{s}$ that may exist in the generated captions $\mathcal{P'}$ with the target attributes $p_{t}$ to make edited prompt $\mathcal{P^{*}}$. As we assume that the conditions that the users give the two words, which correspond to the source and target attributes respectively, for the replacement operation, we intuitively consider injecting source attributes into the generated captions at any index. But, such a simple approach degrades the performance of image editing when the synonym with source attributes exists in the generated captions because both the token of the synonym and target attribute try to edit identical regions. To prevent this editing region collision, we should consider the captions that include synonyms and remove them.

To take into account this problem in the image captioning framework, we propose a simple but effective method as shown in the~\figref{fig:captioning_method}. The captioner $LM(\cdot)$ produces the caption $\mathcal{P'}$ corresponding to the input image. To find the synonym with the source attribute in the $\mathcal{P}'$, we utilize the CLIP's text encoder and calculate the text-text similarity score. We can get the index $k$ that has a Top-1 similarity score. We can assume the token at the index is the synonym of the source attribute $p_{s}$ so we replace the token at index $k$ om $\mathcal{P'}$ into the $p_{s}$. Additionally, we throw tokens away after an index of $k$ to make a truncated prompt $\mathcal{P}_{k}$. To generate a feasible prompt, we generated a new caption conditioned on the $\mathcal{P}_{k}=[p_{1}, \ldots, p_{s}]$ as follows:
\begin{equation}
{p}^{'}_{k+1} = LM( \mathcal{P}_{k} ; E_{I}(\mathcal{I}) )
\end{equation}
The reason for such a process is to utilize the prior pre-trained captioning model to decide whether this replacement operation is reasonable. If we can find the wrong index and replace the token, the captioning model, generate weird captions because it is conditioned on the erroneous prompts.
Finally, to check whether the captioner generates the erroneous caption or not, we compare the image-text CLIP score between two prompts, $\mathcal{P}^{'}_{k}$ and the other one  $\mathcal{P}_{M+1}$ that appends the source attribute $p_{s}$ into the prompts $\mathcal{P'}$ at the location of $M+1$. The prompt that has a higher score is chosen as the final prompt.

\begin{figure}[t]
\centering
\includegraphics[width=0.75\linewidth]{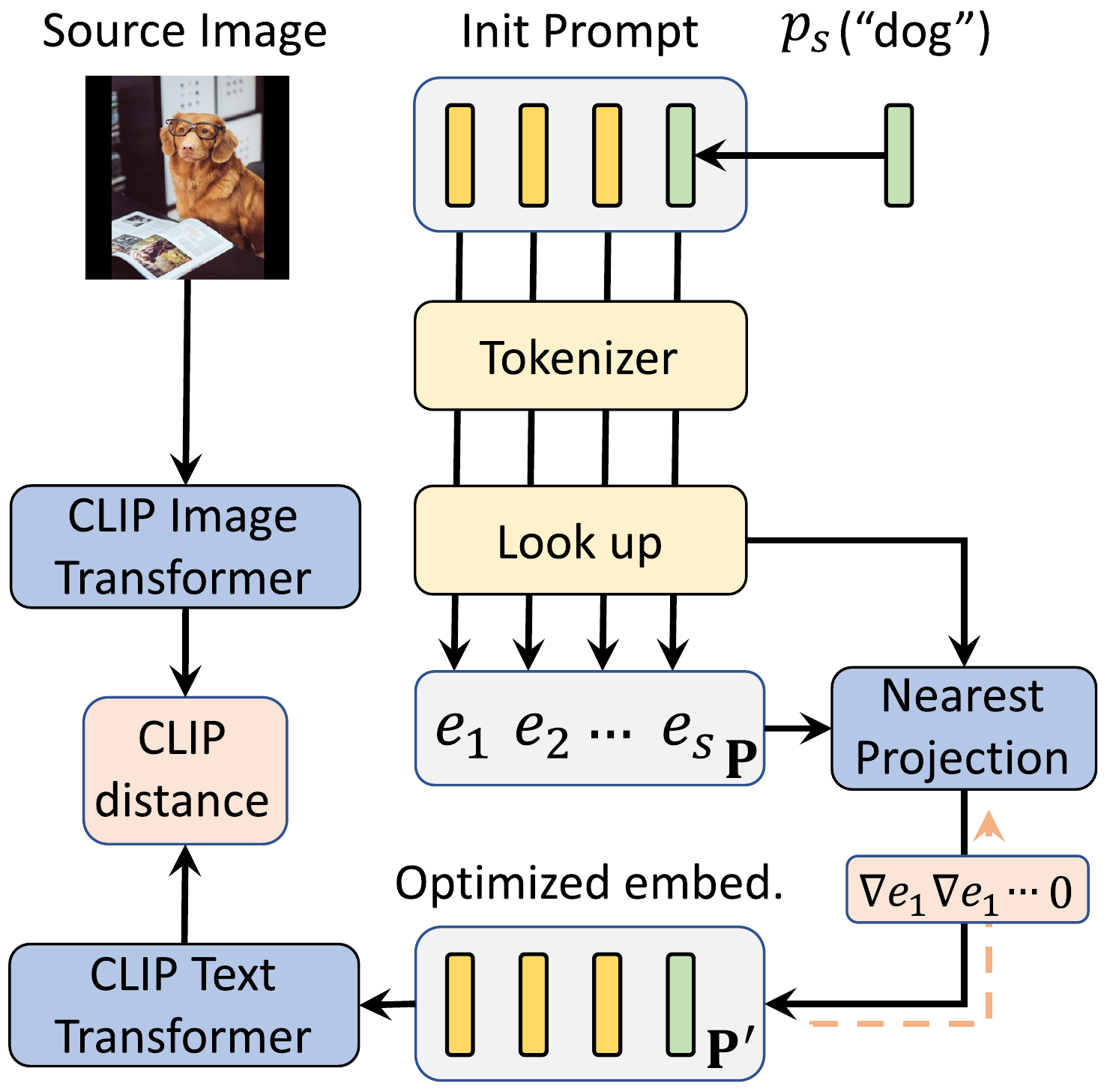}\\
\vspace{-5pt}
    \caption{\textbf{Illustration of optimization method.}   
    }
    \label{fig:optimization_method}\vspace{-10pt}
\end{figure}

\vspace{3pt}
\subsection{Optimization-based method}
In this section, we then propose an optimization-based method.
The PEZ~\cite{wen2023hard} is a training-free hard prompt optimization framework that maximizes the CLIP score on optimized prompts and corresponding images. Given a frozen model text encoder $\psi_{T}(\cdot)$, a sequence of learnable embeddings, $\mathbf{P}=[\mathbf{e_i}, ... \mathbf{e_M}], \mathbf{e_i} \in \mathbb{R}^{d} $, where $M$ is the number of tokens worth of vectors to optimize, and $d$ is the dimension of the embeddings. Additionally, we employ an objective function $\mathcal{L}_\text{CLIP}$:
\begin{equation}
\mathcal{L_{\text{CLIP}}}(P, \mathcal{I}) = {D}( \psi_{T}(\mathbf{P}_{proj}), \psi_{I}(\mathcal{I}) )
\end{equation}
where $\psi_{I}$ denotes the CLIP's image encoder and ${D}$ is the cosine distance function between text and image embedding vectors from CLIP encoders.
The discreteness of the token space comes by using a projection function, $\text{Proj}_{\mathbf{E}}$, that takes the individual embedding vectors $\mathbf{e_i}$ in the prompt and projects them to their nearest neighbor in the embedding look-up matrix $E^{|V|\times d}$ where $|V|$ is the vocabulary size of the model, and we denote the result of this operation as  $\mathbf{P}_\text{proj}=\text{Proj}_{\mathbf{E}}(\mathbf{P}):=[\text{Proj}_{\mathbf{E}}(\mathbf{e_i}), ... \text{Proj}_{\mathbf{E}}(\mathbf{e_M})]$. Finally, the hard prompts learned recurrently following the equation:
\begin{equation}
\mathbf{P} = \mathbf{P} - \gamma\nabla_{\mathbf{P}_\text{proj}} \mathcal{L_{\text{CLIP}}}(\mathbf{P}_\text{proj}, \mathcal{I})
\end{equation}
Finally, the optimized embeddings $P'$ are decoded by tokenizer to $\mathcal{P'}$ that has the highest CLIP score during the optimization process.

For injecting target words, we can not utilize the same solution with the captioning model in this hard prompt optimization framework. Such an optimization process tends to ignore the erroneous tokens and focus on maximizing the CLIP score by using the other tokens. On the contrary, it denotes that if the informative token is located in the prompts, the other tokens focus on searching for other contextual meanings to maximize the CLIP score.

From this assumption, before the optimization process, we experimented to append the tokens of source attributes $p_{s}$ at the randomly initialized prompts. To prevent the token of the source attribute from changing to other tokens, we force the gradient to the token of the source attribute to be zero as shown in the~\figref{fig:optimization_method}.

Although these frameworks generate feasible prompts, sometimes the redundant tokens are captured, degrading the editing performance. To solve this problem, we should remove the redundant tokens and preserve informative tokens. The prompt distillation methods proposed in PEZ~\cite{wen2023hard} can be adopted to reduce the specific number of tokens. However, it needs an additional optimization process for prompt distillation and it spends tens of seconds. Furthermore, it sometimes captures new redundant tokens because it just maximizes the text-text CLIP similarity scores to previously captured prompts. We find that by comparing the image-text similarity between sets of ablated prompts $\mathcal{P}^{abl}_{M}$ and generated prompt $\mathcal{P'}$, where the ablated prompts are made by removing the token at location $m$ in the range of $[0, M]$. If the CLIP score of ablated prompts is higher than generated one, it denotes that the token located at abated index is not informative. As a result, the tokes in that indices are removed and others are preserved. This simple comparison only spends a few seconds to remove redundant tokens and preserve informative tokens.

\begin{figure*}[t]
\centering
  \begin{subfigure}{0.32\linewidth}
  \centering
	{\includegraphics[width=1\linewidth]{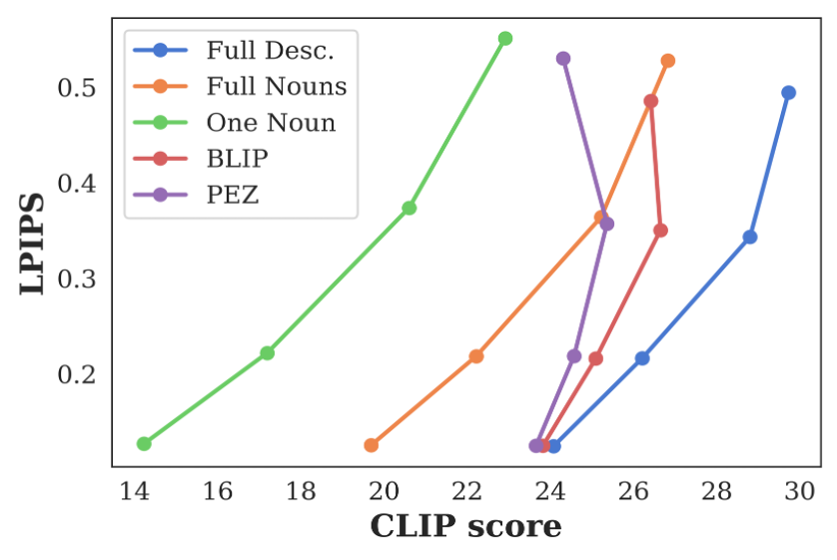}}\caption{\small{SDEdit~\cite{meng2021sdedit}}}\hfill
    \end{subfigure}
  \begin{subfigure}{0.32\linewidth}
  \centering
	{\includegraphics[width=1\linewidth]{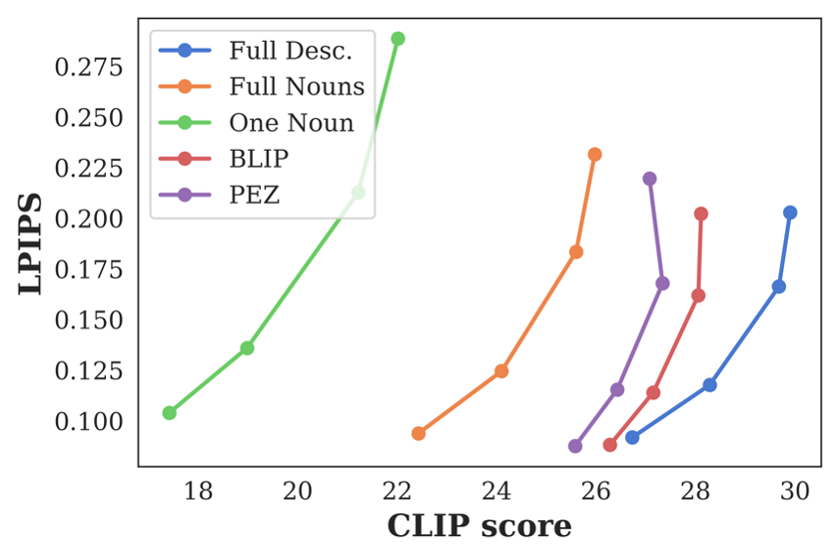}}\caption{\small{Prompt-to-Prompt~\cite{hertz2022prompt}}}\hfill
    \end{subfigure}
  \begin{subfigure}{0.32\linewidth}
  \centering
	{\includegraphics[width=1\linewidth]{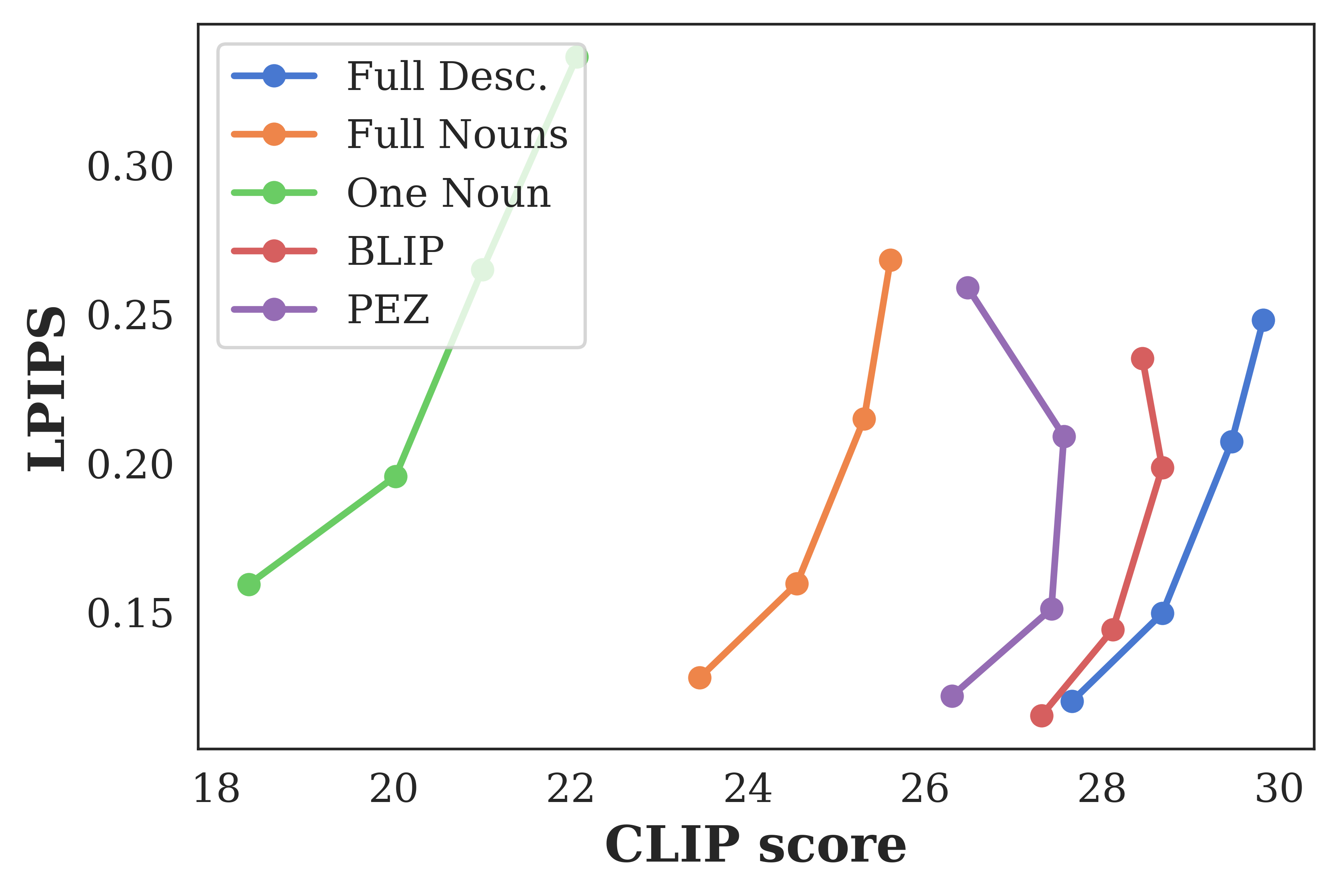}}\caption{\small{Null-text Inversion~\cite{mokady2022null}}}\hfill
    \end{subfigure}

	\vspace{-12pt}
    \caption{\textbf{Quantitative comparison on various methods on  different levels of text-conditionings with generated images.}}
    \label{fig:quan_gen}\vspace{-5pt}
\end{figure*}

\begin{figure*}[t]
\centering
  \begin{subfigure}{0.32\linewidth}
  \centering
	{\includegraphics[width=1\linewidth]{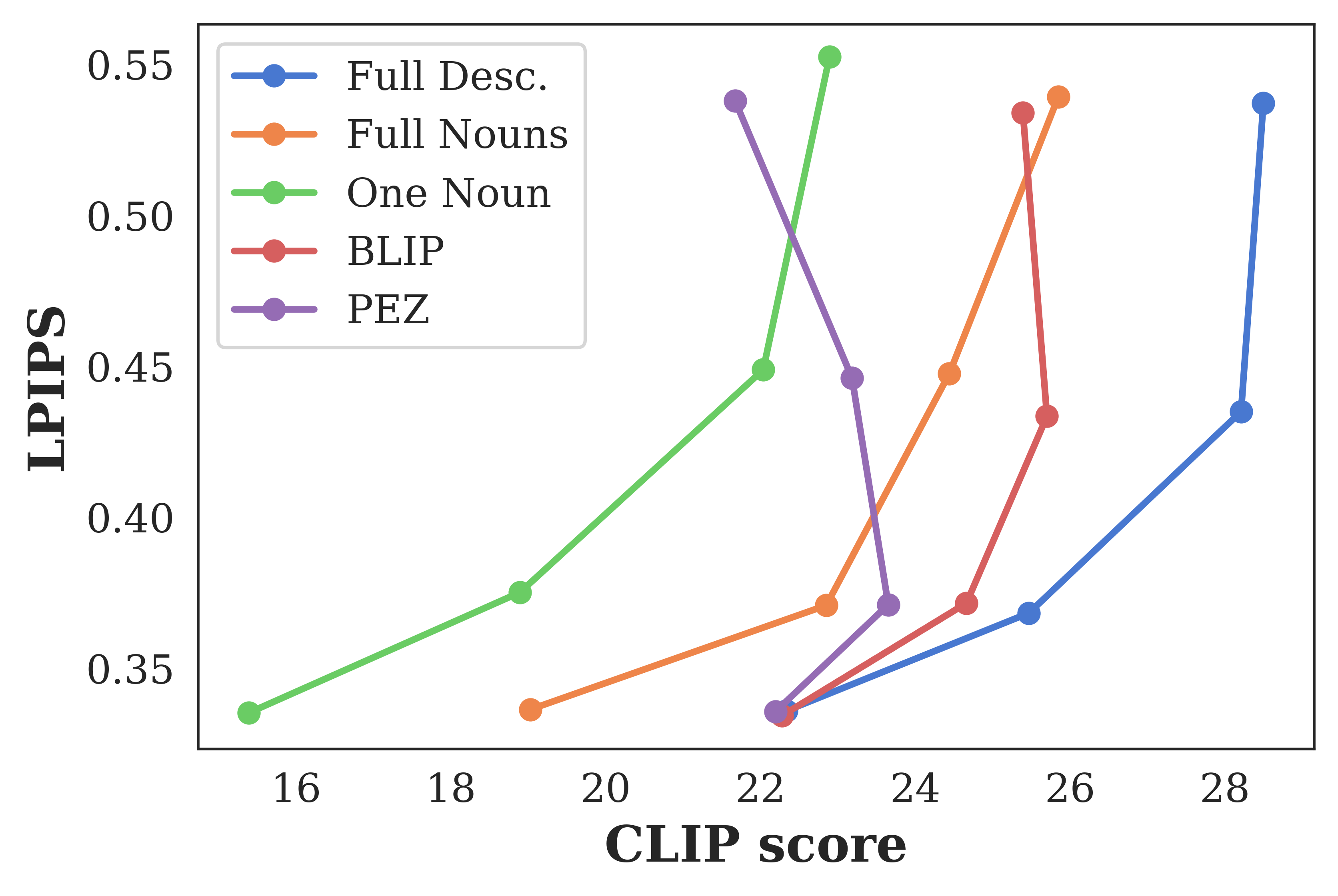}}\caption{\small{SDEdit~\cite{meng2021sdedit}}}\hfill
    \end{subfigure}
  \begin{subfigure}{0.32\linewidth}
  \centering
	{\includegraphics[width=1\linewidth]{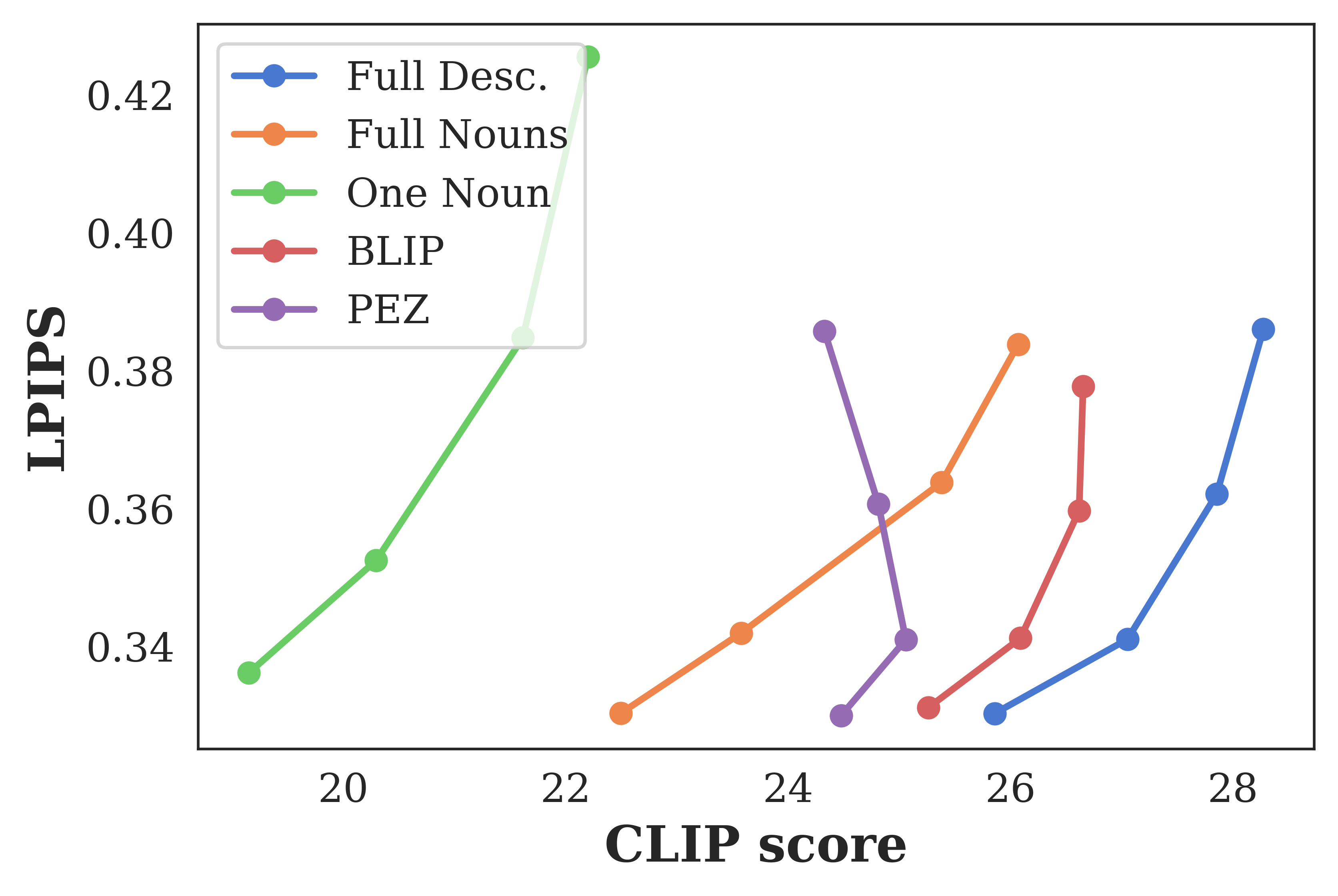}}\caption{\small{Prompt-to-Prompt~\cite{hertz2022prompt}}}\hfill
    \end{subfigure}
  \begin{subfigure}{0.32\linewidth}
  \centering
	{\includegraphics[width=1\linewidth]{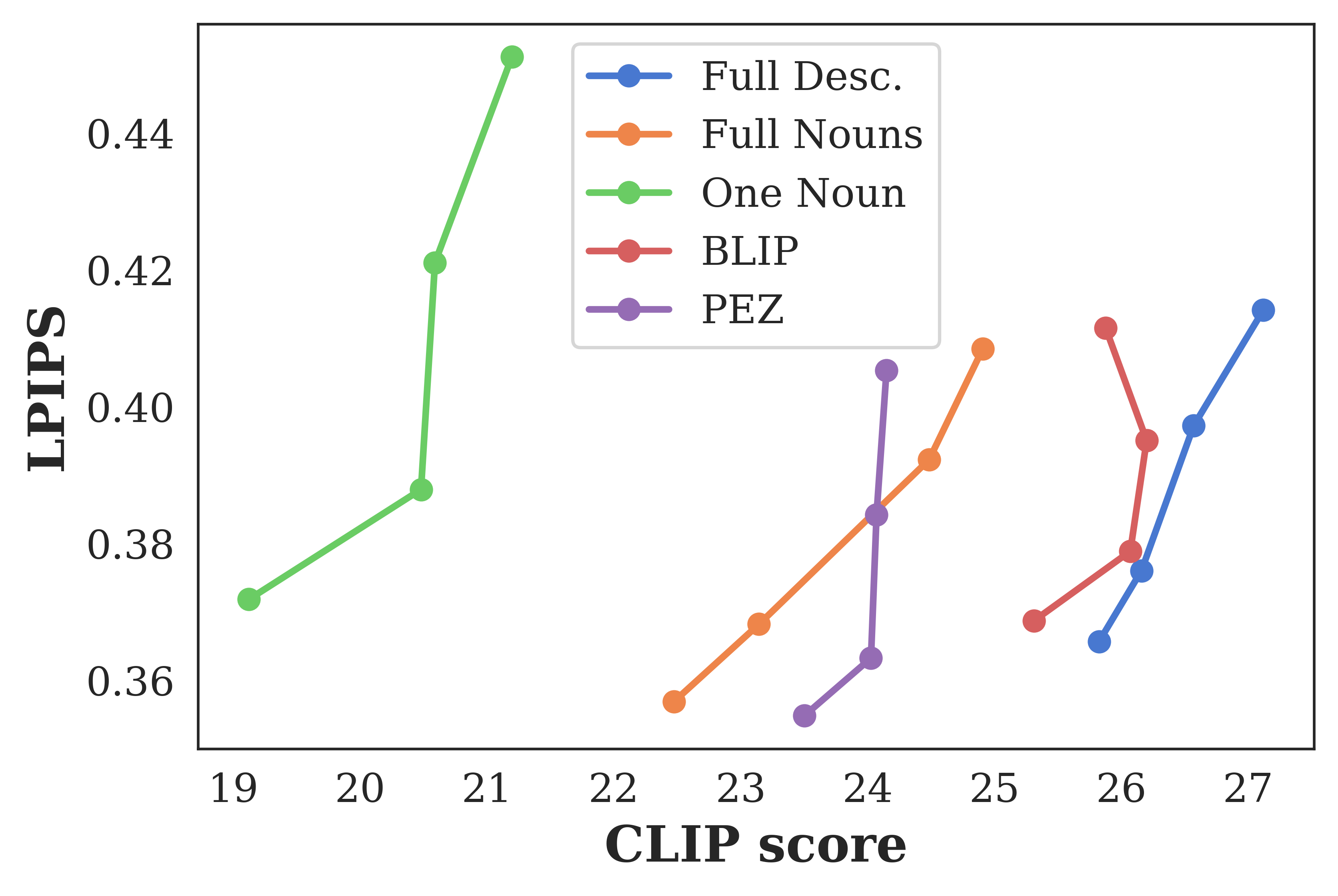}}\caption{\small{Null-text Inversion~\cite{mokady2022null}}}\hfill
    \end{subfigure}

	\vspace{-12pt}
    \caption{\textbf{Quantitative comparison on various methods on different levels of text-conditionings with real images.}}
    \label{fig:quan_real}\vspace{-5pt}
\end{figure*}

\begin{figure}[t]
    \centering
    \setlength{\abovecaptionskip}{0pt}
    \setlength{\belowcaptionskip}{0pt}
    \setlength{\tabcolsep}{0.55pt}

    \begin{tabular}{c c c c c c c}
    \includegraphics[width=0.16\linewidth,height=0.16\linewidth]{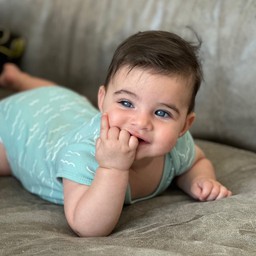} &
    \includegraphics[width=0.16\linewidth,height=0.16\linewidth]{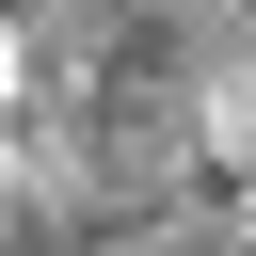} &
    \includegraphics[width=0.16\linewidth,height=0.16\linewidth]{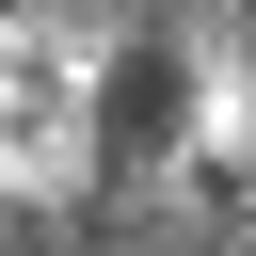} &
    \includegraphics[width=0.16\linewidth,height=0.16\linewidth]{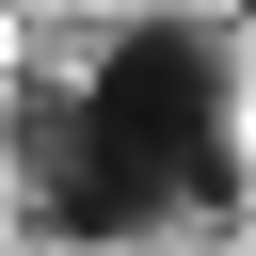} &
    \includegraphics[width=0.16\linewidth,height=0.16\linewidth]{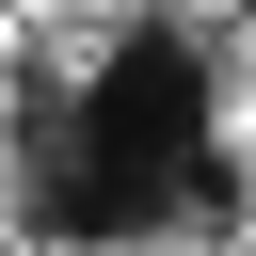} &
    \includegraphics[width=0.16\linewidth,height=0.16\linewidth]{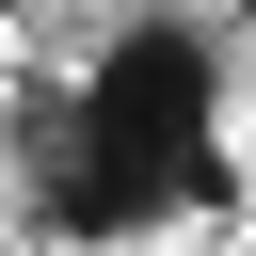} \\

    {\small\begin{tabular}{c@{}c@{}c@{}c@{}c@{}} Input \\Sample\end{tabular}} &
    {\small\begin{tabular}{c@{}c@{}c@{}c@{}c@{}} One \\Noun\end{tabular}} &
    {\small\begin{tabular}{c@{}c@{}c@{}c@{}c@{}} Full \\Nouns\end{tabular}} &
    {\small\begin{tabular}{c@{}c@{}c@{}c@{}c@{}} Full \\Desc.\end{tabular}} &
    {\small\begin{tabular}{c@{}c@{}c@{}c@{}c@{}} PEZ \\ caption\end{tabular}} &
    {\small\begin{tabular}{c@{}c@{}c@{}c@{}c@{}} BLIP \\ caption \end{tabular}}
    \end{tabular}
    
    \vspace{2pt}
    
    \caption{\textbf{Localization capability.} We visualize the cross-attention map for the token of ``sofa''.}
    \label{fig:localization}
     \vspace{-12pt}
\end{figure}

\vspace{5pt}
\section{Experiments}
\label{sec:experiments}
This section provides experimental details and validation of the proposed work. The qualitative and quantitative comparisons are shown in the \secref{comparison}. The~\secref{abl} provides the ablation study about our proposed methods.

\subsection{Experimental settings}
\begin{figure*}[!t]
    \centering
    \setlength{\abovecaptionskip}{6.5pt}
    \setlength{\belowcaptionskip}{-3.5pt}
    \setlength{\tabcolsep}{0.55pt}
    \renewcommand{\arraystretch}{1.0}
    {
    
    \begin{tabular}{c}
    
    \begin{tabular}{c@{\hskip 5pt} c c c @{\hskip 10pt} c c c@{\hskip 10pt} c c c}\\

    \raisebox{0.045\linewidth}{\small\begin{tabular}{c@{}c@{}} Source \end{tabular}} &
    \includegraphics[width=0.09\linewidth,height=0.09\linewidth]{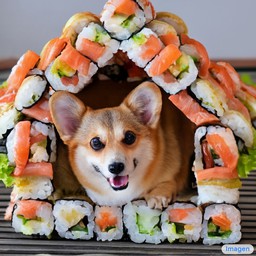} & 
    \includegraphics[width=0.09\linewidth,height=0.09\linewidth]{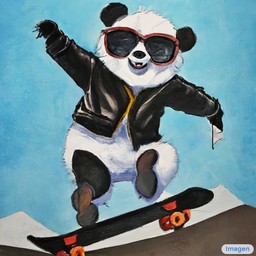} & 
    \includegraphics[width=0.09\linewidth,height=0.09\linewidth]{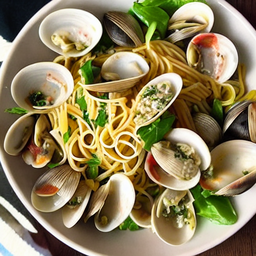}
    \\
    
    \raisebox{0.045\linewidth}{\small\begin{tabular}{c@{}c@{}} One \\ Noun \end{tabular}} &
    \includegraphics[width=0.09\linewidth,height=0.09\linewidth]{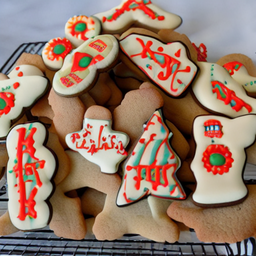} & 
    \includegraphics[width=0.09\linewidth,height=0.09\linewidth]{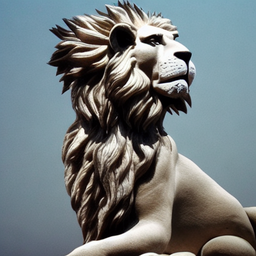} & 
    \includegraphics[width=0.09\linewidth,height=0.09\linewidth]{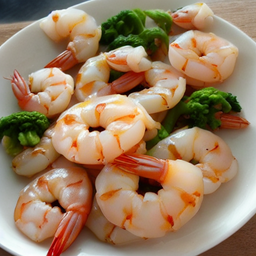} & 
    \includegraphics[width=0.09\linewidth,height=0.09\linewidth]{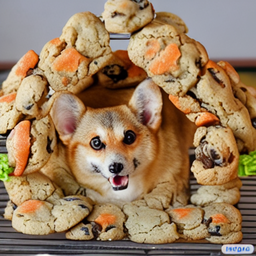} & 
    \includegraphics[width=0.09\linewidth,height=0.09\linewidth]{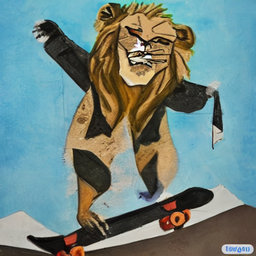} &
    \includegraphics[width=0.09\linewidth,height=0.09\linewidth]{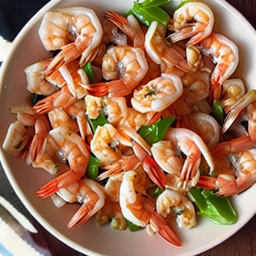} &
    \includegraphics[width=0.09\linewidth,height=0.09\linewidth]{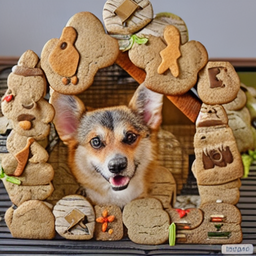} & 
    \includegraphics[width=0.09\linewidth,height=0.09\linewidth]{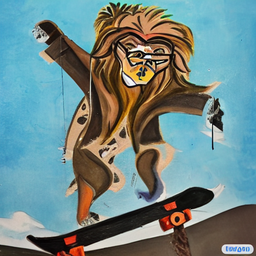} &
    \includegraphics[width=0.09\linewidth,height=0.09\linewidth]{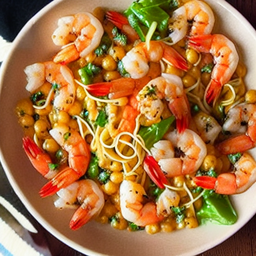}
    \\[-3pt]
    
    \raisebox{0.045\linewidth}{\small\begin{tabular}{c@{}c@{}c@{}c@{}} Full \\ Nouns \end{tabular}}  &
    \includegraphics[width=0.09\linewidth,height=0.09\linewidth]{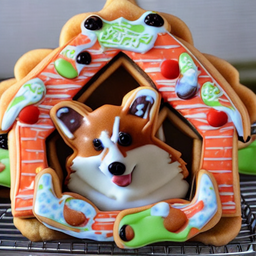} & 
    \includegraphics[width=0.09\linewidth,height=0.09\linewidth]{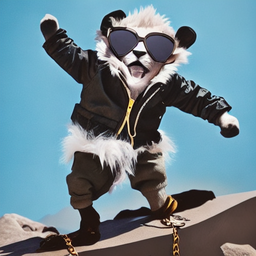} & 
    \includegraphics[width=0.09\linewidth,height=0.09\linewidth]{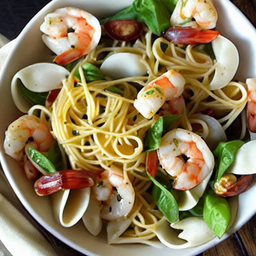} & 
    \includegraphics[width=0.09\linewidth,height=0.09\linewidth]{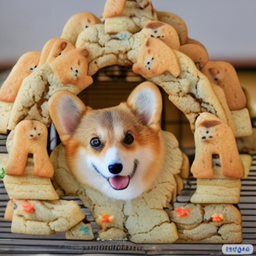} & 
    \includegraphics[width=0.09\linewidth,height=0.09\linewidth]{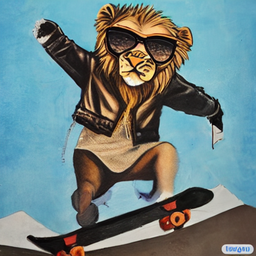} &
    \includegraphics[width=0.09\linewidth,height=0.09\linewidth]{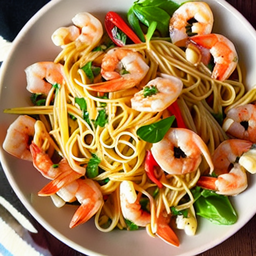} &
    \includegraphics[width=0.09\linewidth,height=0.09\linewidth]{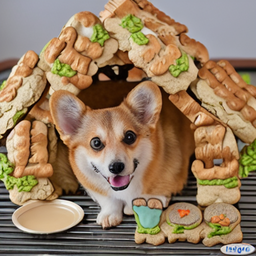} & 
    \includegraphics[width=0.09\linewidth,height=0.09\linewidth]{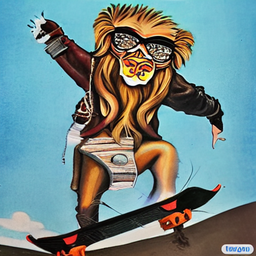} & 
    \includegraphics[width=0.09\linewidth,height=0.09\linewidth]{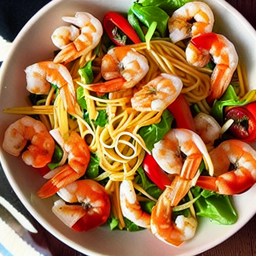}
    \\[-3pt]
    
    \raisebox{0.045\linewidth}{\small\begin{tabular}{c@{}c@{}c@{}c@{}} Full \\ Descriptions \end{tabular}}  &
    \includegraphics[width=0.09\linewidth,height=0.09\linewidth]{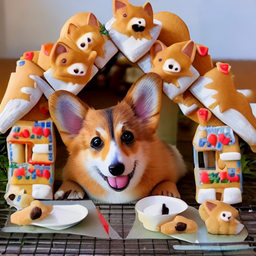} & 
    \includegraphics[width=0.09\linewidth,height=0.09\linewidth]{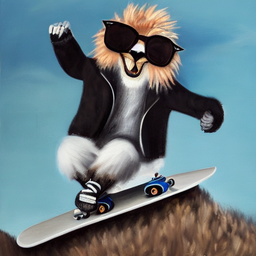} & 
    \includegraphics[width=0.09\linewidth,height=0.09\linewidth]{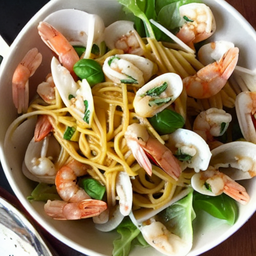} & 
    \includegraphics[width=0.09\linewidth,height=0.09\linewidth]{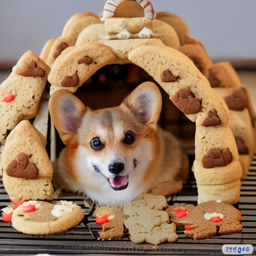} & 
    \includegraphics[width=0.09\linewidth,height=0.09\linewidth]{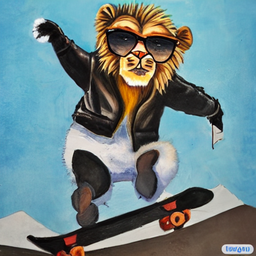} &
    \includegraphics[width=0.09\linewidth,height=0.09\linewidth]{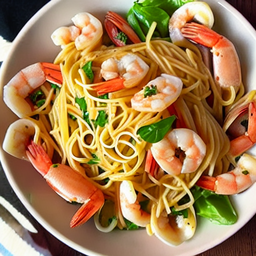} & 
    \includegraphics[width=0.09\linewidth,height=0.09\linewidth]{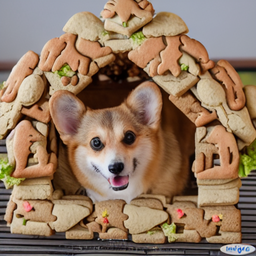} & 
    \includegraphics[width=0.09\linewidth,height=0.09\linewidth]{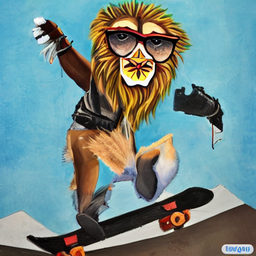} & 
    \includegraphics[width=0.09\linewidth,height=0.09\linewidth]{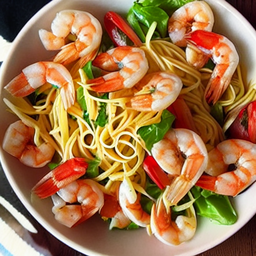}
    \\[-2pt]

    \raisebox{0.045\linewidth}{\small\begin{tabular}{c@{}c@{}c@{}c@{}} PEZ-based\\caption \end{tabular}}  &
    \includegraphics[width=0.09\linewidth,height=0.09\linewidth]{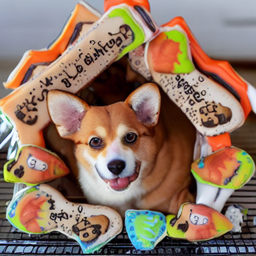} & 
    \includegraphics[width=0.09\linewidth,height=0.09\linewidth]{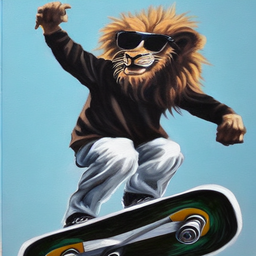} & 
    \includegraphics[width=0.09\linewidth,height=0.09\linewidth]{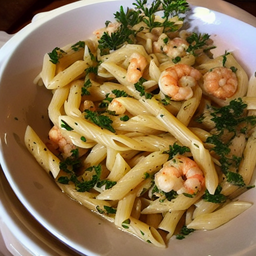} & 
    \includegraphics[width=0.09\linewidth,height=0.09\linewidth]{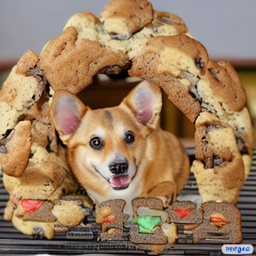}& 
    \includegraphics[width=0.09\linewidth,height=0.09\linewidth]{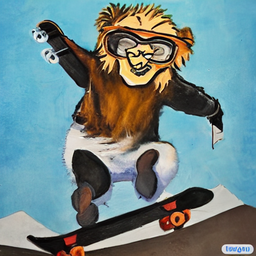} &
    \includegraphics[width=0.09\linewidth,height=0.09\linewidth]{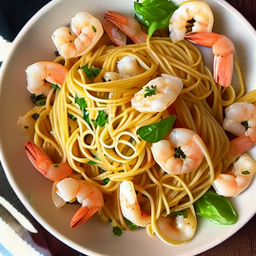} & 
    \includegraphics[width=0.09\linewidth,height=0.09\linewidth]{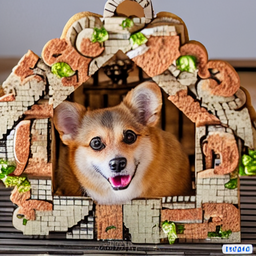} & 
    \includegraphics[width=0.09\linewidth,height=0.09\linewidth]{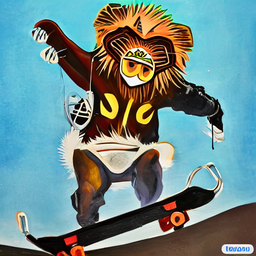} & 
    \includegraphics[width=0.09\linewidth,height=0.09\linewidth]{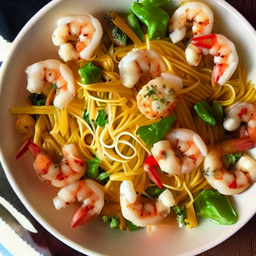}
    \\[-2pt]

    \raisebox{0.045\linewidth}{\small\begin{tabular}{c@{}c@{}c@{}c@{}} BLIP-based\\caption \end{tabular}}  &
    \includegraphics[width=0.09\linewidth,height=0.09\linewidth]{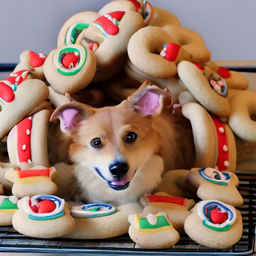} & 
    \includegraphics[width=0.09\linewidth,height=0.09\linewidth]{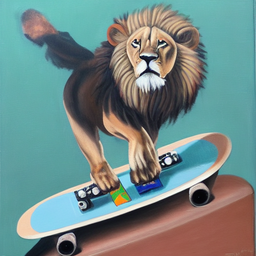} & 
    \includegraphics[width=0.09\linewidth,height=0.09\linewidth]{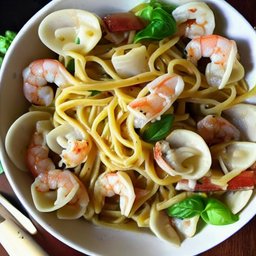} & 
    \includegraphics[width=0.09\linewidth,height=0.09\linewidth]{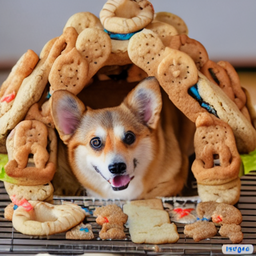}& 
    \includegraphics[width=0.09\linewidth,height=0.09\linewidth]{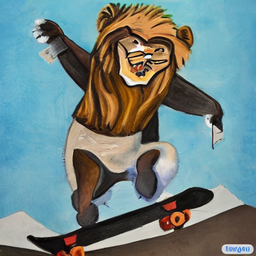} &
    \includegraphics[width=0.09\linewidth,height=0.09\linewidth]{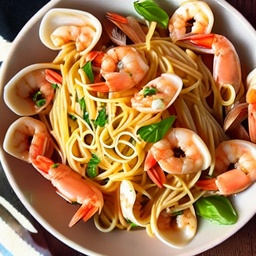} & 
    \includegraphics[width=0.09\linewidth,height=0.09\linewidth]{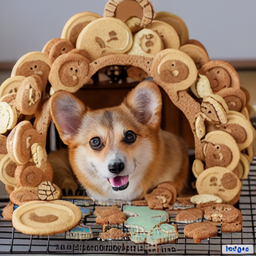} & 
    \includegraphics[width=0.09\linewidth,height=0.09\linewidth]{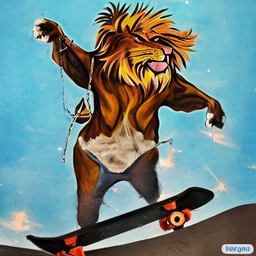} & 
    \includegraphics[width=0.09\linewidth,height=0.09\linewidth]{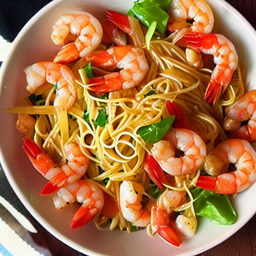}
    \\[-2pt]
    
    \multicolumn{1}{c}{} & \multicolumn{3}{c}{\small{SDEdit}~\cite{meng2021sdedit}} & 
    \multicolumn{3}{c}{\small{Prompt-to-Prompt}~\cite{hertz2022prompt}} &
    \multicolumn{3}{c}{\small{Null-text Inversion}~\cite{mokady2022null}}
    \\[5pt]

    \end{tabular}
    
    \end{tabular}
    }
    \vspace{-8pt}
    \caption{\textbf{Qualitative results on the generated images in the different levels of text conditions.} Given images and target text conditions, we show the different quality of editing results with different levels of text conditions on each model respectively. 
    }
    \label{fig:qual_gen}
\end{figure*}

Evaluation is provided in~\cref{fig:qual_real,fig:qual_gen,fig:quan_real,fig:quan_gen}. We evaluated these on the real and synthesized images which are collected from the MSCOCO~\cite{caesar2018coco} and other works~\cite{mokady2022null, bar2022text2live, couairon2022diffedit, hertz2022prompt}. In total, we use 50 samples of images, source captions, and edited captions. We then applied all the editing frameworks~\cite{meng2021sdedit,hertz2022prompt,mokady2022null} on each image-caption pair using the default Stable Diffusion~\cite{rombach2021highresolution} hyper-parameters for an increasing number of iterations per diffusion step.

\vspace{-9pt}
\paragraph{Baseline editing models.}
We use latent diffusion models~\cite{rombach2022high}. We use the 890M parameter text-conditional model trained on LAION-5B~\cite{schuhmann2021laion}, known as \textit{Stable Diffusion}, at $512\!\times\!512$ resolution.
Since these models operate in VQGAN latent spaces~\cite{esser2021taming}, the resolution is $64\!\times\!64$. We use 50 steps in DDIM sampling with a fixed schedule on the Prompt-to-Prompt~\cite{hertz2022prompt}, DiffEdit~\cite{couairon2022diffedit}. On the other hand, we search for the optimal $t$ step for evaluating SDEdit~\cite{meng2021sdedit}. We also use classifier-free guidance \cite{ho2022classifier} with the recommended value of 7.5 for Stable Diffusion.

\vspace{12pt}
\subsection{Comparisons}
~\label{comparison}
Our approach focuses on utilizing the generated prompts for image editing to give an intuitive experience to the users, so we compare among different prompts settings, which are ``One Noun'', ``Full Nouns'', ``Full Description'', PEZ-based captions, and BLIP-based captions, on the SDEdit~\cite{meng2021sdedit}, Prompt-to-Prompt~\cite{hertz2022prompt} and Null-text Inversion~\cite{mokady2022null}. In total, we use 50 samples of images, source captions, and edited captions. In addition, particularly, we compare the PEZ- and BLIP-based captions which produce prompts that have different characteristics. Lastly, we consider the concurrent work of Instruct-pix2-pix~\cite{brooks2022instructpix2pix} and pix2pix-zero~\cite{parmar2023zero}, which employs GPT-3~\cite{brown2020language} to serve the intuitive editing framework to users.

\vspace{-8pt}
\paragraph{Quantitative results}
We report the quantitative comparison among different levels of prompts, PEZ~\cite{wen2023hard} and BLIP~\cite{li2022blip} based methods on the editing methods, SDEdit~\cite{meng2021sdedit}, Prompt-to-Prompt~\cite{hertz2022prompt}, Null text Inversion~\cite{mokady2022null}. We adopt the two metrics to evaluate text-image correspondence with CLIP score~\cite{radford2021learning} and perceptual similarity between original and edited images by using LPIPS~\cite{zhang2018perceptual}. As shown in the~\cref{fig:quan_real,fig:quan_gen}, in most cases, the full description setting shows the highest scores in both CLIP score and LPIPS, and the performance of BLIP- and PEZ- based captions overwhelm the performance of the one noun setting. Although all the cases have source attributes and corresponding target attributes in the prompts, the observed performance gap is significant. We can find an interesting point about the PEZ- and BLIP-based captions at the right-up side corner of the graph. Although such methods produce informative captions that ground source images well, they sometimes act as redundant information in image editing. From the observation, we can infer that other informative tokens, except for attributes to be edited, affect the editing performance. Such an analysis is naturally connected to the~\figref{fig:localization} which shows localization capability among the different levels of prompt settings. Consequently, this proves that inaccurate localization quality is one of the causes of performance degradation.

\begin{figure*}[!t]
    \centering
    \vspace{-12pt}
    \setlength{\abovecaptionskip}{6.5pt}
    \setlength{\belowcaptionskip}{-3.5pt}
    \setlength{\tabcolsep}{0.55pt}
    \renewcommand{\arraystretch}{1.0}
    {
    
    \begin{tabular}{c}
    
    \begin{tabular}{c@{\hskip 5pt} c c c @{\hskip 10pt} c c c@{\hskip 10pt} c c c}\\

    \raisebox{0.045\linewidth}{\small\begin{tabular}{c@{}c@{}} Source \end{tabular}} &
    \includegraphics[width=0.09\linewidth,height=0.09\linewidth]{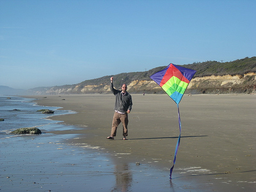} & 
    \includegraphics[width=0.09\linewidth,height=0.09\linewidth]{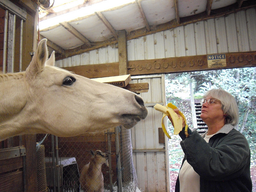} & 
    \includegraphics[width=0.09\linewidth,height=0.09\linewidth]{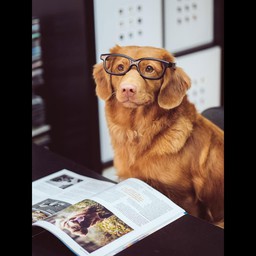} \\
    
    \raisebox{0.045\linewidth}{\small\begin{tabular}{c@{}c@{}} One \\ Noun \end{tabular}} &
    \includegraphics[width=0.09\linewidth,height=0.09\linewidth]{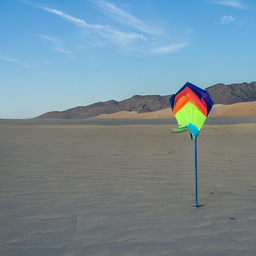} & 
    \includegraphics[width=0.09\linewidth,height=0.09\linewidth]{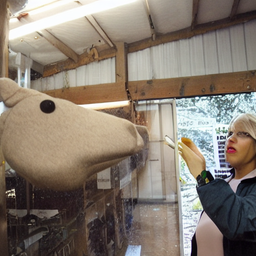} & 
    \includegraphics[width=0.09\linewidth,height=0.09\linewidth]{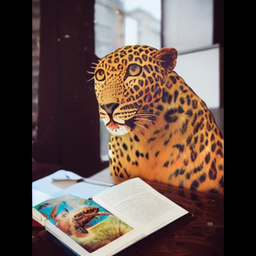} & 
    \includegraphics[width=0.09\linewidth,height=0.09\linewidth]{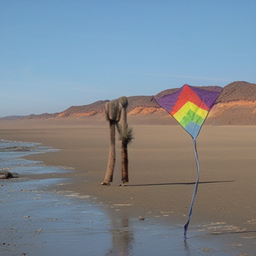} & 
    \includegraphics[width=0.09\linewidth,height=0.09\linewidth]{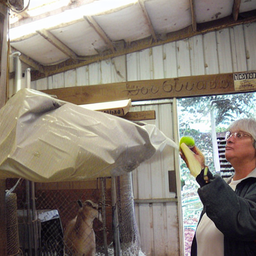} &
    \includegraphics[width=0.09\linewidth,height=0.09\linewidth]{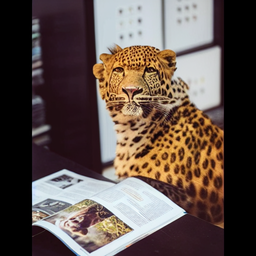} &
    \includegraphics[width=0.09\linewidth,height=0.09\linewidth]{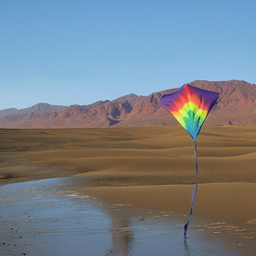} & 
    \includegraphics[width=0.09\linewidth,height=0.09\linewidth]{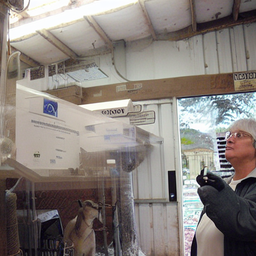} & 
    \includegraphics[width=0.09\linewidth,height=0.09\linewidth]{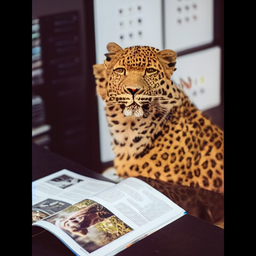} \\[-3pt]
    
    \raisebox{0.045\linewidth}{\small\begin{tabular}{c@{}c@{}c@{}c@{}} Full \\ Nouns \end{tabular}}  &
    \includegraphics[width=0.09\linewidth,height=0.09\linewidth]{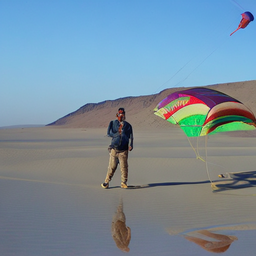} & 
    \includegraphics[width=0.09\linewidth,height=0.09\linewidth]{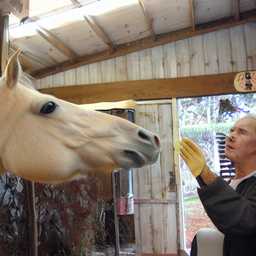} & 
    \includegraphics[width=0.09\linewidth,height=0.09\linewidth]{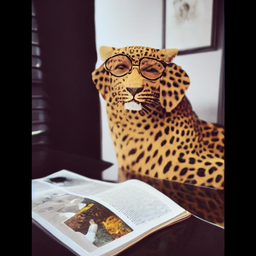} & 
    \includegraphics[width=0.09\linewidth,height=0.09\linewidth]{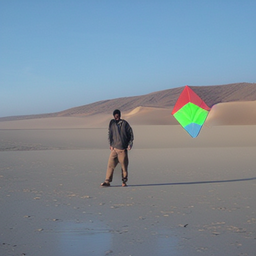} & 
    \includegraphics[width=0.09\linewidth,height=0.09\linewidth]{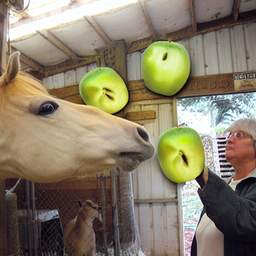} &
    \includegraphics[width=0.09\linewidth,height=0.09\linewidth]{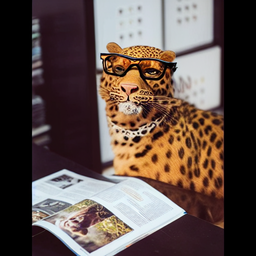} &
    \includegraphics[width=0.09\linewidth,height=0.09\linewidth]{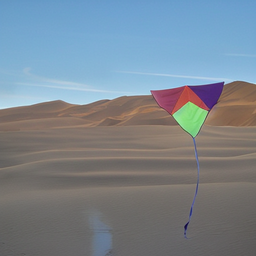} & 
    \includegraphics[width=0.09\linewidth,height=0.09\linewidth]{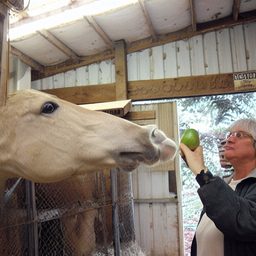} & 
    \includegraphics[width=0.09\linewidth,height=0.09\linewidth]{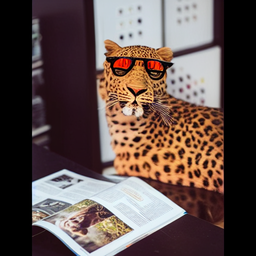} \\[-3pt]
    
    \raisebox{0.045\linewidth}{\small\begin{tabular}{c@{}c@{}c@{}c@{}} Full \\ Descriptions \end{tabular}}  &
    \includegraphics[width=0.09\linewidth,height=0.09\linewidth]{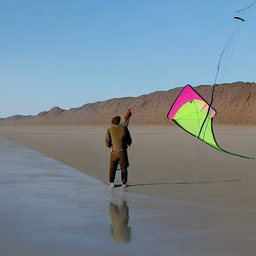} & 
    \includegraphics[width=0.09\linewidth,height=0.09\linewidth]{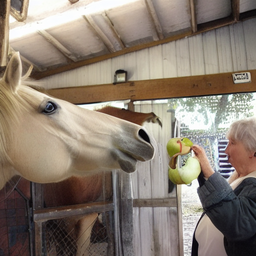} & 
    \includegraphics[width=0.09\linewidth,height=0.09\linewidth]{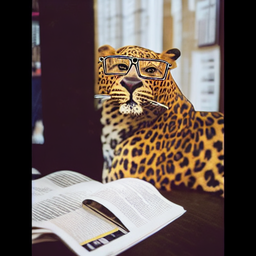} & 
    \includegraphics[width=0.09\linewidth,height=0.09\linewidth]{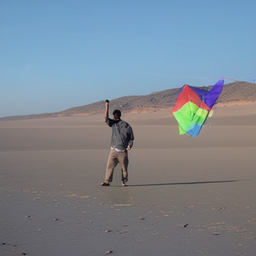} & 
    \includegraphics[width=0.09\linewidth,height=0.09\linewidth]{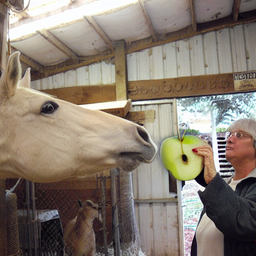} &
    \includegraphics[width=0.09\linewidth,height=0.09\linewidth]{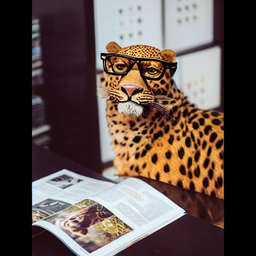} & 
    \includegraphics[width=0.09\linewidth,height=0.09\linewidth]{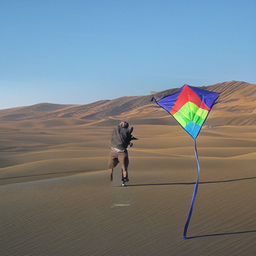} & 
    \includegraphics[width=0.09\linewidth,height=0.09\linewidth]{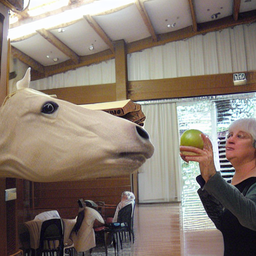} & 
    \includegraphics[width=0.09\linewidth,height=0.09\linewidth]{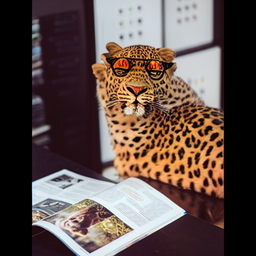} \\[-2pt]

    \raisebox{0.045\linewidth}{\small\begin{tabular}{c@{}c@{}c@{}c@{}} PEZ-based \\ caption \end{tabular}}  &
    \includegraphics[width=0.09\linewidth,height=0.09\linewidth]{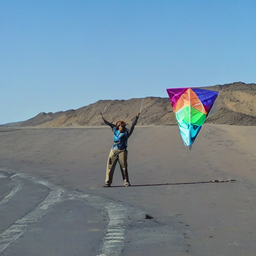} & 
    \includegraphics[width=0.09\linewidth,height=0.09\linewidth]{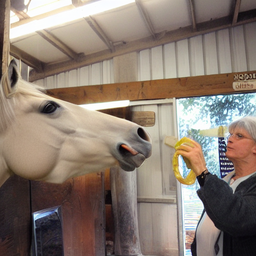} & 
    \includegraphics[width=0.09\linewidth,height=0.09\linewidth]{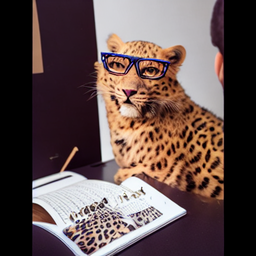} & 
    \includegraphics[width=0.09\linewidth,height=0.09\linewidth]{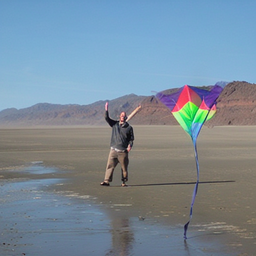}& 
    \includegraphics[width=0.09\linewidth,height=0.09\linewidth]{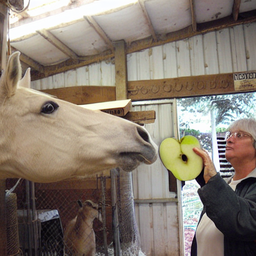} &
    \includegraphics[width=0.09\linewidth,height=0.09\linewidth]{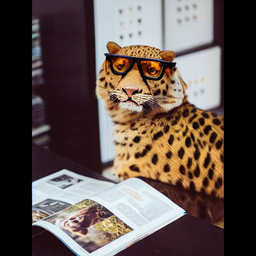} & 
    \includegraphics[width=0.09\linewidth,height=0.09\linewidth]{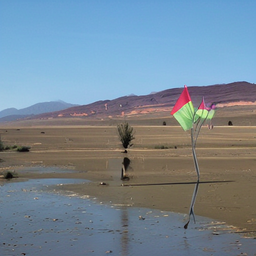} & 
    \includegraphics[width=0.09\linewidth,height=0.09\linewidth]{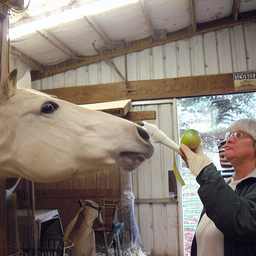} & 
    \includegraphics[width=0.09\linewidth,height=0.09\linewidth]{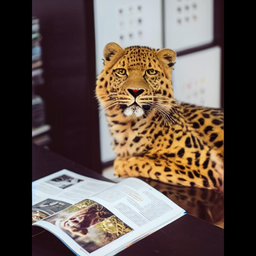} \\[-2pt]

    \raisebox{0.045\linewidth}{\small\begin{tabular}{c@{}c@{}c@{}c@{}} BLIP-based \\ caption \end{tabular}}  &
    \includegraphics[width=0.09\linewidth,height=0.09\linewidth]{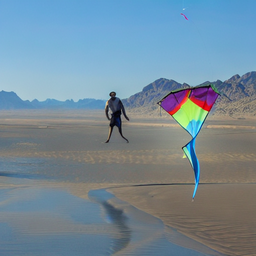} & 
    \includegraphics[width=0.09\linewidth,height=0.09\linewidth]{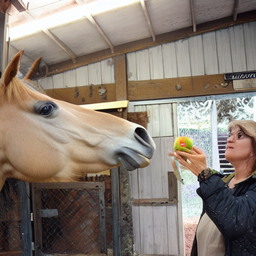} & 
    \includegraphics[width=0.09\linewidth,height=0.09\linewidth]{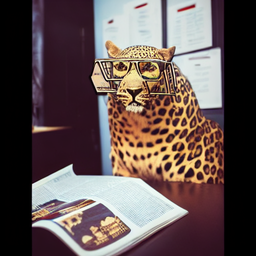} & 
    \includegraphics[width=0.09\linewidth,height=0.09\linewidth]{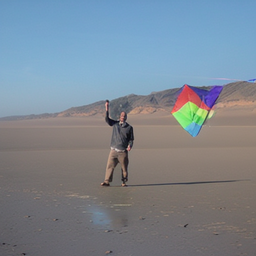}& 
    \includegraphics[width=0.09\linewidth,height=0.09\linewidth]{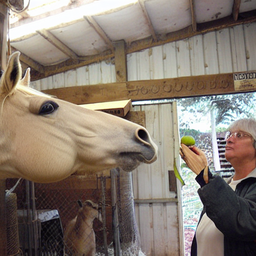} &
    \includegraphics[width=0.09\linewidth,height=0.09\linewidth]{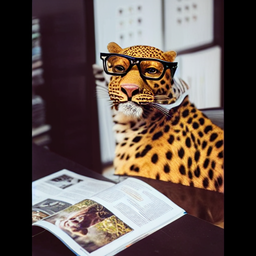} & 
    \includegraphics[width=0.09\linewidth,height=0.09\linewidth]{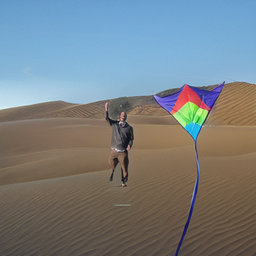} & 
    \includegraphics[width=0.09\linewidth,height=0.09\linewidth]{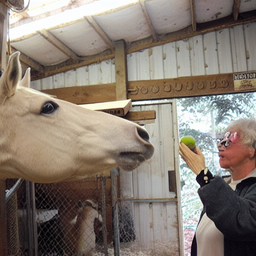} & 
    \includegraphics[width=0.09\linewidth,height=0.09\linewidth]{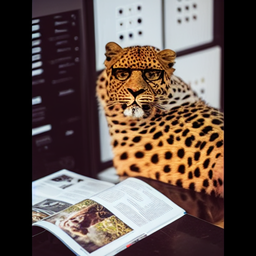} \\[-2pt]
    
 \multicolumn{1}{c}{} & \multicolumn{3}{c}{\small{SDEdit}~\cite{meng2021sdedit}} & 
 \multicolumn{3}{c}{\small{Prompt-to-Prompt}~\cite{hertz2022prompt}} &
 \multicolumn{3}{c}{\small{Null-text Inversion}~\cite{mokady2022null}}  \\[5pt]

    \end{tabular}
    
    \end{tabular}
    }
    \vspace{-7pt}
    \caption{\textbf{Qualitative results on the real images in the different levels of text conditions.} Given images and target text conditions, we show the different quality of editing results with different levels of text conditions on each model respectively. 
    }
    \label{fig:qual_real}
\end{figure*}

\vspace{-7pt}
\paragraph{Qualitative results}
We show visual results for~\cref{fig:quan_real,fig:quan_gen} in~\cref{fig:qual_real,fig:qual_gen}, respectively. We observe that the quality differences among different levels of prompts, full description setting, PEZ~\cite{wen2023hard} and BLIP~\cite{li2022blip} based methods are not significant. In addition, we can find the emerging problems that we discussed in~\secref{analysis}. The edited images by only using one noun tend to suffer more quality degradation than other settings. For example, in the~\cref{fig:qual_gen}, the one noun setting loose other contextual information such as ``corgi'' and ``sunglasses''. Interestingly, we can also find that the performance gap between one noun and full nouns settings is larger than between full nouns and full descriptions settings. It indicates that nouns are more critical than verbal words and more informative when we edit images with text conditions. In addition, as shown in the last row in the~\figref{fig:qual_gen}, the BLIP-based caption failed to capture the word ``sunglasses'' and produced the results without sunglasses which is not unintended.  Especially, the edited results by utilizing generated prompts are competitive with the Full Description setting which means the target attributes injected into the generated prompts and target regions are successfully edited.

\vspace{-7pt}
\paragraph{Comparison between PEZ- and BLIP-based editing}
From~\cref{fig:qual_real,fig:qual_gen},~\cref{fig:quan_real,fig:quan_gen}, in most of the cases, the BLIP-based editing shows higher performance than PEZ-based editing results. Such phenomena result from the property of each framework. The PEZ~\cite{wen2023hard} is designed to optimize each text token by maximizing the image-text CLIP score. The generated hard prompts are not a form of the sentence, but rather close to the set of nouns, such as ``daughters rowland pino percy lovely horses moment seaside fra''. However, the image captioning method in BLIP~\cite{li2022blip} generates the prompts in the form of a sentence, such as ``two children riding horses on the beach''. Although these two prompts ground the same number of objects, the existence of verbal words such as ``riding'' affects the performance. Furthermore, when we calculate the image-text CLIP score with these prompts, the prompts from the PEZ usually show higher CLIP scores than ones from BLIP. It means that the prompts which have high image-text CLIP could indicate good descriptions for the corresponding image, but do not always assure it gives better editing results.  

\begin{table}
    \centering
    \scalebox{0.9}{
    \begin{tabular}{l c c}\toprule
        Method                    & CLIP Score($\uparrow$) &LPIPS($\downarrow$)\\\midrule
        Instruct pix2pix~\cite{brooks2022instructpix2pix} &\textbf{31.54}&{0.298}  \\ 
        pix2pix-zero~\cite{parmar2023zero}                &\textbf{32.86}&{0.367}\\ 
        PEZ-based caption                                 &{27.13}&\textbf{0.25}\\ 
        BLIP-based caption                                &{28.71}&\textbf{0.25} \\ \bottomrule
    \end{tabular}
    }
    \vspace{-1mm}
    \caption{\textbf{Comparison to other instructive editing frameworks.}}
    \label{tab:comp_pix2pix}
    \vspace{-10pt}
\end{table}

\begin{table*}
    \centering
    \scalebox{.9}{
    \begin{tabular}{l c c c c c c }\toprule
         &\multicolumn{2}{c}{Start}
         &\multicolumn{2}{c}{Middle}
         &\multicolumn{2}{c}{End}
         \\
         \cmidrule(lr){2-3} \cmidrule(lr){4-5} \cmidrule(lr){6-7}
        Method  & CLIP Score($\uparrow$) &LPIPS($\downarrow$) & CLIP Score($\uparrow$) &LPIPS($\downarrow$)   & CLIP Score($\uparrow$) &LPIPS($\downarrow$) \\\midrule 
        PEZ-based caption  &{25.66}&{0.300}  &{25.73}&\textbf{0.283}  &\textbf{26.21}&{0.288}   \\ 
        BLIP-based caption &{27.09}&{0.297}  &{28.05}&{0.298}  &\textbf{28.07}&\textbf{0.292}\\ \bottomrule
    \end{tabular}
    }
    \vspace{-1mm}
    \caption{\textbf{Editing performance according to different injection locations of target words.}}
    \label{tab:loc}
\end{table*}

\begin{table*}
    \centering
    \scalebox{.9}{
    \begin{tabular}{l c c c c c c c c}\toprule
         &\multicolumn{2}{c}{4}
         &\multicolumn{2}{c}{8}
         &\multicolumn{2}{c}{16}
         \\
         \cmidrule(lr){2-3} \cmidrule(lr){4-5} \cmidrule(lr){6-7}
        Method                    & CLIP Score($\uparrow$) &LPIPS($\downarrow$) & CLIP Score($\uparrow$) &LPIPS($\downarrow$)   & CLIP Score($\uparrow$) &LPIPS($\downarrow$) \\\midrule 
        PEZ-based caption
        &\textbf{26.21}&\textbf{0.288}  &{25.63}&{0.293}  &{24.89}&0.303\\ 
        BLIP-based caption
        &{27.37}&{0.298}  &\textbf{28.07}&{0.292}  &{26.73}&\textbf{0.281}\\ \bottomrule
    \end{tabular}
    }
    \vspace{-1mm}
    \caption{\textbf{Ablation study on the number of utilized tokens.}}
    \label{tab:num_token}
\end{table*}

\vspace{-6pt}
\paragraph{Comparison to other instructive frameworks}
We conduct additional experiments to compare to other instructive text-driven editing frameworks, such as Instruct pix2pix~\cite{brooks2022instructpix2pix} and pix2pix-zero~\cite{parmar2023zero}. The Instruct pix2pix utilize the cross-attention swapping technique that is proposed in the Prompt-to-Prompt to collect paired dataset to train the instructive diffusion model. Meanwhile, the Pix2pix-zero constrains the structure information by guiding the difference of outputs between the source image and edited image in the cross-attention layer. Hence, we compare editing results on Prompt-to-Prompt to these works in the~\tabref{tab:comp_pix2pix}. We can observe that editing results from the generated captions are competitive to the Instruct-pix2pix and show higher performance than pix2pix-zero. The reason is that the pix2pix-zero utilized the generated caption from BLIP as the text condition and it sometimes does not include the source attributes to be edited. 

\vspace{-4pt}
\subsection{Ablation study}
~\label{abl}
We conduct ablation studies about injecting location, number of tokens, and techniques for removing redundant text tokens. We experiment on all the editing frameworks in~\figref{fig:qual_real} and utilize data in~\cref {fig:quan_real,fig:quan_gen}. All the metrics in the~\cref {tab:loc,tab:num_token,tab:token_abl} are calculated respectively and averaged.

\paragraph{Injection locations of source attributes}
In~\tabref{tab:loc}, we validate the effects of the injection locations of source attributes in the generated prompts. For validation, we can find the performance gap between the three settings is considerably high. However, injecting the source attribute at the end of the generated prompts shows the highest score, so we set the end of the location at the generated prompts as the default injecting location.

\vspace{-7pt}
\paragraph{Number of tokens}
The hyper-parameter about the number of tokens is important because shorter prompts do not have sufficient semantics, on the other hand, the longer prompts tend to capture redundant tokens. So we ablate the optimal number of tokens. In~\tabref{tab:num_token}, we can observe that a length of 4 has the best score in both metrics on PEZ captions. In, BLIP-based captions, the length of 8 has the best score but, has a lower LPIPS score than the length of 16. Considering the trade-off between CLIP score and LPIPS, we set the default value in BLIP-based caption as 8.

\paragraph{Removing and preserving tokens}
To determine which token is informative or redundant is important. To validate the proposed method, we compare the two cases when we apply the prompt distillation technique~\cite{wen2023hard} and the proposed technique. \tabref{tab:token_abl} shows that our proposed method more effectively removes and preserves tokens.

\begin{table}
    \centering
    \scalebox{0.7}{
    \begin{tabular}{l c cc c c c}\toprule
         &\multicolumn{2}{c}{Prompt Distill.}
         &\multicolumn{2}{c}{Prompt Abl.(Ours)}
         \\
         \cmidrule(lr){2-3} \cmidrule(lr){4-5} \cmidrule(lr){6-7}
        Method                & CLIP Score($\uparrow$) &LPIPS($\downarrow$) & CLIP Score($\uparrow$) &LPIPS($\downarrow$)\\\midrule 
        PEZ-based     &{26.51}&{0.299} &\textbf{26.30}&\textbf{0.295}  \\ 
        BLIP-based    &{27.65}&{0.299} &\textbf{28.13}&\textbf{0.290} \\ \bottomrule
    \end{tabular}
    }
    \vspace{-1mm}
    \caption{\textbf{Removing and preserving tokens.}}
    \label{tab:token_abl}
    \vspace{-15pt}
\end{table}

\section{Conclusions}
In this paper, we conduct a comprehensive analysis of the impact of source prompts on image editing performance. To this end, we categorize prompts into three levels based on their degree of information in prompts: ``One Noun'', ``Full Nouns'', and ``Full Description''. Our findings suggest that using only one noun as a prompt for image editing leads to a decrease in performance, despite users preferring simpler prompts. To address this problem, we propose a method for preventing the user's prompt engineering process by combining the state-of-the-art prompt generation frameworks. In detail, as we can not directly utilize prompts from these frameworks, we devise the scheme for injecting source attributes that we want to edit in the input image into the generated prompt. In addition, we propose a simple and fast technique to remove the redundant and preserve informative text tokens.

For future work, we aim to explore metrics that can help determine the most suitable prompts for image editing, as well as extend our approach to recently proposed other editing frameworks and image captioning models.

{\small
\bibliographystyle{ieee_fullname}
\bibliography{egbib}
}

\end{document}

%% file: title.tex
\twocolumn[
{
\renewcommand\twocolumn[1][]{#1}%
\maketitle
\begin{center}
    \centering
    \captionsetup{type=figure}
\includegraphics[width=1.0\textwidth,height=6cm]{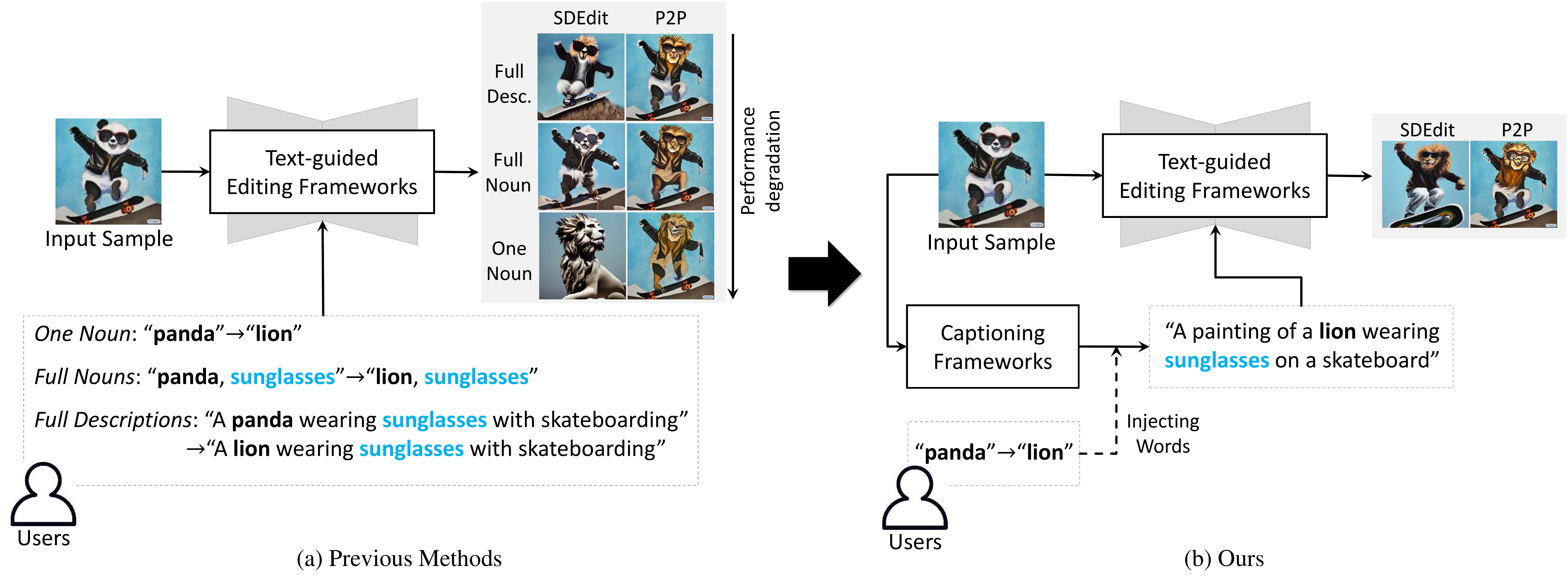}
    \vspace{-15pt}
    \captionof{figure}{\textbf{Illustration of problems in existing text-guided editing frameworks and our proposed method.} When utilizing the existing text-guided editing frameworks~\cite{meng2021sdedit,hertz2022prompt,mokady2022null}, the users should consider which prompts are better conditions to edit input images. However, it is time-consuming and makes erroneous text conditions which cause performance degradation. To address these problems, we first divide and define the level of text conditions. Then we propose a method that effectively utilizes the existing captioning frameworks to grounding input images while the users only give simple words to edit the target region.}
    \label{fig:teaser}\vspace{12pt}
\end{center}%
}]
\thispagestyle{empty}